\documentclass[twocolumn,letterpaper]{article}

\PassOptionsToPackage{table}{xcolor}
\usepackage[numbers,compress]{natbib}
\usepackage{paper-theme} 
\usepackage{fontawesome5}
\usepackage{subcaption}

\usepackage[most]{tcolorbox}   % 'most' needed for enhanced + breakable

\definecolor{tableblue}{HTML}{E8F1F7}

\newtcolorbox[auto counter]{takeaway}[1][]{
  enhanced, breakable,
  colback=tableblue,          % your existing colour
  colframe=tableblue,         % same -> no visible border
  boxrule=0pt, arc=2pt,
  left=6pt, right=6pt, top=5pt, bottom=5pt,
  fonttitle=\bfseries, coltitle=black,
  attach title to upper={\ },
  title={Takeaway~\thetcbcounter:},
  #1
}

\fancypagestyle{fancy}{%
  \fancyhf{}%
  \fancyfoot[C]{\thepage}%
}
\fancypagestyle{plain}{%
  \fancyhf{}%
  \fancyfoot[C]{\thepage}%
}
\papertitle{LeVJEPA: Efficient \& Scalable Video Pretraining\\ without the Heuristics}
\paperauthors{%
  \textbf{Lukas Kuhn}\textsuperscript{1,2}\quad
  \textbf{Lucas Maes}\textsuperscript{4,5}\quad
  \textbf{Giuseppe Serra}\textsuperscript{1,2}\quad
  \textbf{Quentin Le Lidec}\textsuperscript{8}\quad \\
  \textbf{Yann LeCun}\textsuperscript{7,8}\quad
  \textbf{Randall Balestriero}\textsuperscript{\textbf{*},6,8}\quad
  \textbf{Florian Buettner}\textsuperscript{\textbf{*},1,2,3}
}
\paperaffiliations{%
  \textsuperscript{1}German Cancer Research Center\quad
  \textsuperscript{2}German Cancer Consortium\quad
  \textsuperscript{3}Goethe University Frankfurt\quad
  \textsuperscript{4}Mila\quad \\
  \textsuperscript{5}Universit\'e de Montr\'eal\quad
  \textsuperscript{6}Brown University\quad
  \textsuperscript{7}Courant Institute, New York University\quad \\
  \textsuperscript{8}Advanced Machine Intelligence (AMI Labs)\quad

}
\papercontribution{\textsuperscript{*}Equal advising.\quad Correspondence: \texttt{lukas.kuhn@dkfz-heidelberg.de}}

\begin{document}

\setcounter{topnumber}{2}
\setcounter{totalnumber}{3}
\renewcommand{\topfraction}{0.9}
\renewcommand{\textfraction}{0.1}
\renewcommand{\floatpagefraction}{0.8}
\renewcommand{\dbltopfraction}{0.9}
\renewcommand{\dblfloatpagefraction}{0.8}

% Title and abstract span both columns (passed to \PaperMakeTitle's optional arg).
\PaperMakeTitle[{%
\vspace*{-1.5em}
\begin{paperabstract}

Video carries the temporal structure of the physical world, yet learning representations from it has remained computationally expensive: prevailing self-supervised methods either prevent representation collapse through architectural asymmetries, coupling an exponential-moving-average target encoder, a stop-gradient, and a capacity-limited predictor, or circumvent it by reconstructing masked content in pixel space with a dedicated decoder. We introduce \textbf{LeVJEPA}, the first video encoder trained under LeJEPA's collapse-free objective, which dispenses with both. A single encoder is trained with an invariance loss over global and local views of a clip, regularized by SIGReg, which excludes collapse with a provable guarantee. The trainable architecture consequently reduces to an encoder and a projector, and the objective simplifies to a single hyperparameter. This formulation admits two properties that we examine in this paper. First, the cost of pretraining is governed by the number of tokens the encoder observes; uniform random token dropping renders this number small and, remarkably, simultaneously improves downstream accuracy. At matched epochs on identical data, LeVJEPA matches or surpasses V-JEPA~2 across ViT-S/B/L at $5.6$ to $20.8\times$ less total pretraining compute, and at matched total FLOPs it exceeds the strongest video baseline by $7.6$ points on ImageNet-1K while remaining competitive on motion-centric benchmarks. Second, since no asymmetry between branches is required, the encoder can be trained with block-causal attention at no measurable accuracy cost, such that the representation of every frame is a function of past observations alone: temporal ordering becomes a property of the encoder itself. Against a compute-matched DINOv2 trained on frames of the same videos, LeVJEPA approaches the image-pretrained encoder on appearance-centric evaluation while nearly doubling its motion-centric accuracy. These results indicate that, once its computational overhead is removed, video becomes a viable and in several respects preferable substrate for general-purpose visual pretraining.

\end{paperabstract}
\vspace{10pt}
\begin{center}
\begin{minipage}{0.98\textwidth}
  \captionsetup{type=figure}%
  \centering
  \includegraphics[width=\linewidth]{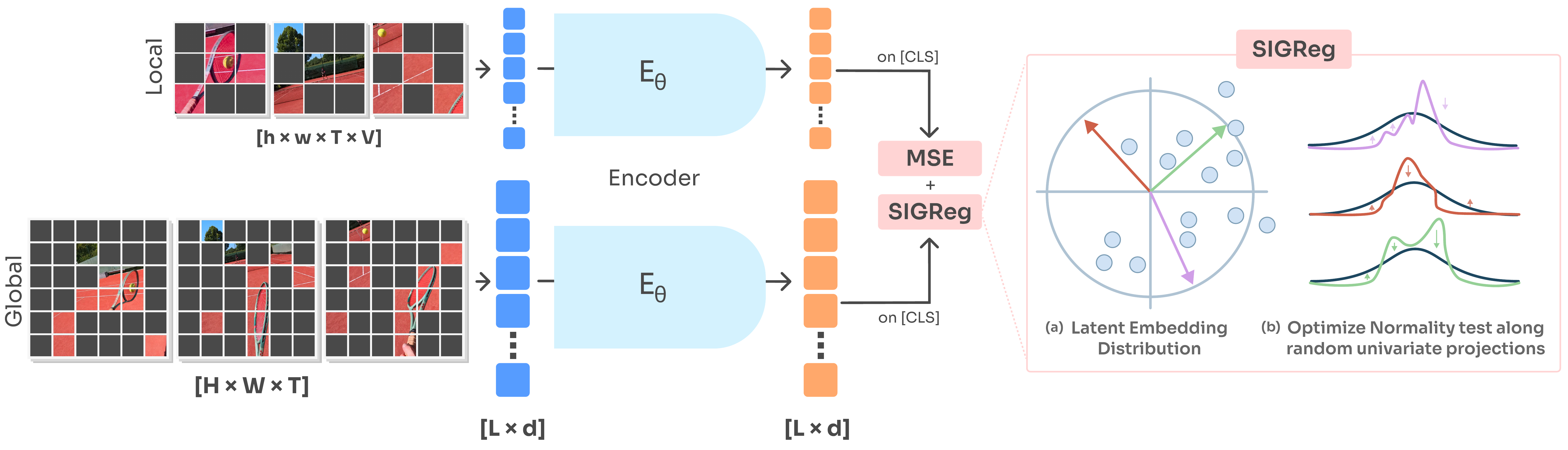}
  \caption{\textbf{LeVJEPA Training.} From each clip we construct one global view ($H \times W \times T$) and several local views ($h \times w \times T \times V$), all sharing the same temporal window; local views are additionally cropped and photometrically augmented. Within every view, $95\%$ of the patch tokens are dropped uniformly at random (grey), and only the retained tokens enter the token sequence ($L \times d$). Both views are processed by the same shared block-causal encoder $E_\theta$, and from the output tokens ($L \times d$) the loss reads only the \texttt{[cls]} embedding of each view. The objective combines a mean squared error that drives each local \texttt{[cls]} embedding toward the global one, with gradients flowing through both branches, and SIGReg~\citep{balestriero2025lejepa} \emph{(right)}, which projects the batch of embeddings onto random directions~(a) and penalizes the deviation of each projection from a standard Gaussian via a normality test~(b), constraining the embedding distribution to an isotropic Gaussian and thereby provably excluding collapse. No target encoder, predictor, stop-gradient, or masked-token reconstruction is required. }
  \label{fig:teaser}
\end{minipage}
\end{center}
}]

\section{Introduction}
\label{sec:intro}

Video is the natural substrate for learning representations of the physical world: it is abundant, requires no annotation, and carries the temporal structure (motion, causality, object permanence) that static images cannot supply~\citep{spelke1995spatiotemporal, rao1999predictive}. Yet video has remained the expensive path to visual representations. A single clip comprises an order of magnitude more tokens than an image, and the methods that learn from it have compounded this cost with architectural machinery: leading video joint-embedding methods train two encoders and a predictor to avoid representation collapse~\citep{grill2020bootstrap, bardes2024vjepa}, and inherit masking schemes designed around that machinery rather than around video itself~\citep{tong2022videomae}. In this work, we train a video encoder with SIGReg~\citep{balestriero2025lejepa}, a statistically principled regularizer that excludes collapse with a provable guarantee. This makes previous collapse-prevention, and the design conventions built around it, unnecessary, and we explore in this paper the substantial simplifications and efficiency gains this enables.

Under this formulation, pretraining reduces to a single encoder trained with a single loss: an invariance term between a global view and several local views of a clip, and SIGReg applied to their embeddings. Where prior methods devote a large share of each training step to components that exist only to stabilize learning --- a target-encoder forward pass over the full clip, a predictor over masked queries~\citep{bardes2024vjepa} --- for us every operation contributes directly to the objective, and the cost of a step is governed only by the number of tokens the encoder observes. This admits two consequences that we develop in this paper. First, the observed token set becomes a free parameter of the method rather than a component of a prediction task, and can be made extremely sparse; the resulting encoder attains accuracy comparable to state-of-the-art video joint-embedding methods at a fraction of their pretraining FLOPs. Second, because no asymmetry between branches is required, the attention topology of the encoder is unconstrained, and we exploit this freedom to train with block-causal attention, bidirectional within a frame and causal across frames. Causality aligns the encoder with the predictive feature principle~\citep{rao1999predictive}, which posits that representations of temporally adjacent stimuli be predictive of one another and therefore requires frame representations computable from past observations alone. It is equally consequential at inference: because past frames need not be re-encoded when new frames arrive, the representation of a video can be extended frame by frame at constant incremental cost, a property required by autoregressive world models and streaming settings that bidirectional encoders can only approximate by re-encoding or by fitting a separate temporal model after pretraining~\citep{zhou2025dino, assran2025vjepa2, maes2026leworldmodel}.

We instantiate this formulation as \textbf{LeVJEPA}\footnote{Code and models are available at \url{https://levjepa.github.io/}}, the first video encoder trained under LeJEPA's collapse-free objective, and evaluate it under frozen probing against video and image pretraining baselines retrained on identical data, in both epoch-matched and FLOP-matched regimes. Our experiments support three findings:

\begin{itemize}
    \item \textit{FLOP-efficient video pretraining.} At matched epochs on identical data, LeVJEPA attains accuracy comparable to or exceeding V-JEPA~2 across ViT-S/B/L at $5.6$ to $20.8\times$ less total pretraining compute. Granted an equal total FLOP budget, it leads the strongest video baseline by $7.6$ points on ImageNet-1K, attains the highest Kinetics-400 accuracy, and remains competitive on Something-Something-v2. Against a DINOv2 baseline trained on frames of the same videos at equal compute, LeVJEPA approaches the image-pretrained encoder on appearance-centric evaluation while nearly doubling its motion-centric accuracy. The efficiency extends to accessibility: a ViT-Tiny pretrained for $12$ hours on a single consumer GPU on unlabeled walking videos attains non-trivial ImageNet accuracy.
    \item \textit{Causal frame representations at no accuracy cost.} Block-causal attention matches fully bidirectional attention under frozen probing, so temporal ordering becomes a property of the encoder itself: frame representations are computable from past observations alone and extend to incoming frames without re-encoding, as autoregressive world modeling and streaming inference require, without a separate temporal model fitted after pretraining.
    \item \textit{Significant simplification of the video pretraining recipe.} The trainable architecture reduces to an encoder and a small projector, with no predictor network, target encoder, or stop-gradient, and the objective carries a single hyperparameter that is fixed to its published default in every experiment. The remaining design choices simplify with it: uniform random token dropping replaces structured masking and acts as an augmentation rather than an approximation, raising ImageNet accuracy monotonically from $33.9\%$ when every token is processed to $47.6\%$ when $95\%$ are discarded; temporal patch aggregation at the input is unnecessary, so tokenization is per frame; and dense, semantically organized patch representations emerge although only the clip-level token is supervised, without the auxiliary patch-level objectives of prior work.
\end{itemize}

\begin{figure}[t]
  \centering
  \includegraphics[width=\linewidth]{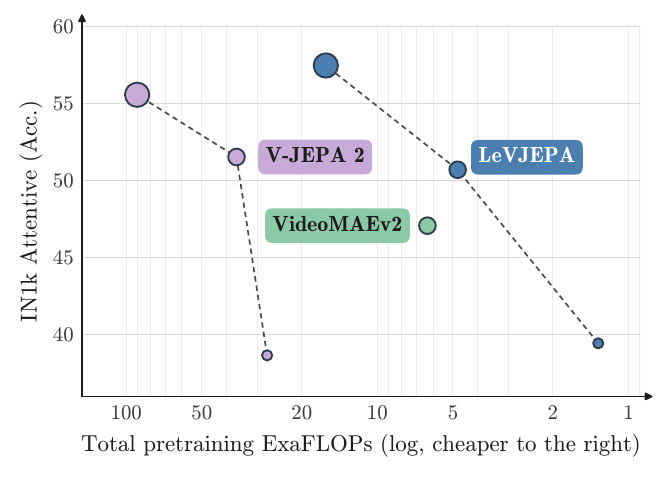}
  \caption{\textbf{Accuracy against total pretraining compute at matched epochs.} ImageNet-1K attentive-probing accuracy of ViT-S, ViT-B, and ViT-L encoders pretrained for $240$ epochs on the identical $20\%$ subsample of K710 (marker size indicates model size; the horizontal axis is logarithmic and reversed).}
  \label{fig:inet_flops}
\end{figure}

\begin{figure*}[t]
  \centering
  \includegraphics[width=0.94\linewidth]{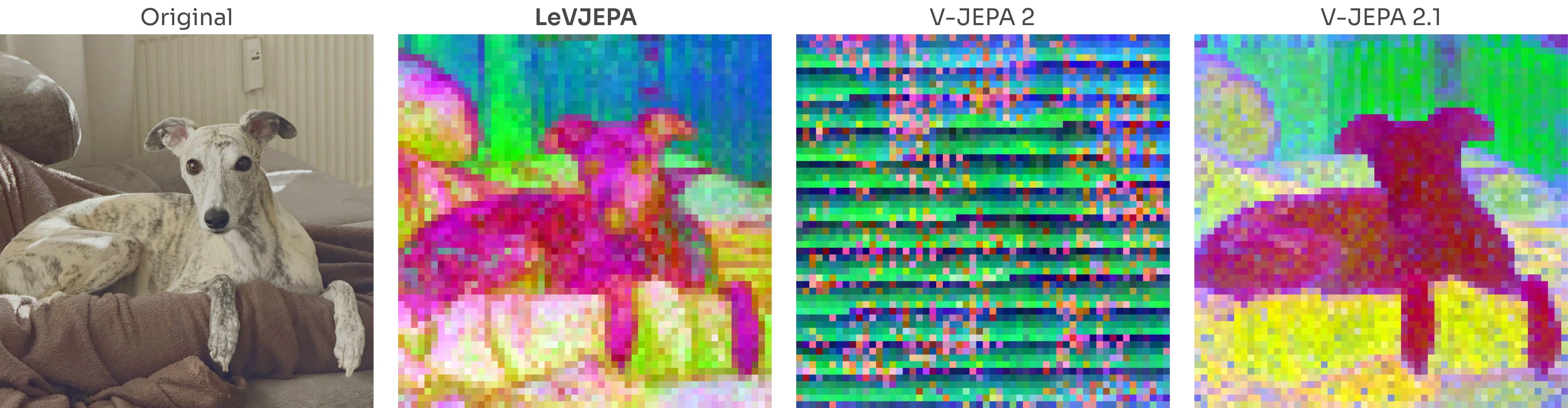}
  \caption{\textbf{Patch-token PCA across methods.} The three leading principal components of the patch-token representations, visualized as RGB, for the same input image. LeVJEPA yields a decomposition that cleanly separates the animal from the surrounding furniture and background, comparable to that of V-JEPA~2.1, which obtains its dense structure through an explicitly introduced auxiliary patch-level loss~\citep{murlabadia2026vjepa21unlockingdense}. V-JEPA~2, trained without such a loss, exhibits no comparable token-level organization. In LeVJEPA, this structure emerges although the training objective supervises only the \texttt{[cls]} token and no patch-level supervision is applied.}
  \label{fig:pca_comparison}
\end{figure*}

\section{Related Work}
\label{sec:related}

\paragraph{Self-supervised learning from images.} Joint-embedding methods map augmented views of the same image to nearby representations, and differ principally in how they exclude the constant solution. BYOL~\citep{grill2020bootstrap} showed that an asymmetry between an online branch and an EMA target branch, with a stop-gradient and a predictor head, suffices without negative pairs. DINO~\citep{caron2021emerging} adopts the same teacher--student arrangement with centering and sharpening, and introduces the multi-crop scheme in which several aggressively cropped local views are matched to a small number of global views. DINOv2~\citep{oquab2024dinov2} combines this with masked-image modelling~\citep{zhou2022image} and scales it on curated data, while I-JEPA~\citep{assran2023self} replaces hand-designed augmentations with prediction between masked regions in representation space. LeVJEPA adopts the global--local view construction of this line, but none of its collapse-prevention machinery.

\paragraph{Explicit constraints on the embedding distribution.} A second line replaces architectural asymmetries with a criterion applied to the embeddings themselves. VICReg~\citep{bardes2021vicreg} and Barlow Twins~\citep{zbontar2021barlow} pair an invariance term with variance and covariance criteria, requiring neither target network nor stop-gradient, but constraining only second-order statistics and saying nothing about which embedding distribution is preferable. LeJEPA~\citep{balestriero2025lejepa} supplies that characterization: under mild assumptions the isotropic Gaussian uniquely minimizes worst-case downstream probing risk, and SIGReg enforces it by reducing the high-dimensional constraint, via the Cram\'er--Wold theorem, to univariate goodness-of-fit tests along random directions at linear cost. The recipe uses a single network and a single loss, and has been validated on static images.

\paragraph{Pixel reconstruction from video.} Masked autoencoding reconstructs held-out content in pixel space and admits no trivial solution, so it needs no collapse prevention. VideoMAE~\citep{tong2022videomae} observes that the temporal redundancy of natural video makes naive masking too easy, since content masked at one instant can be copied from adjacent frames; tube masking, which occludes the same spatial region in every frame, removes this shortcut and permits masking ratios of $90$--$95\%$. VideoMAEv2~\citep{wang2023videomae} scales the approach with an additional decoder-side mask. The tube pattern is therefore a property of the imputation task rather than of video itself, a distinction that becomes visible when the objective imputes nothing (Section~\ref{sec:design}).

\paragraph{Feature prediction and video world models.} V-JEPA~\citep{bardes2024vjepa} predicts representations rather than pixels, training an encoder jointly with a narrow predictor against targets from an exponential-moving-average copy of the encoder; following VideoMAE, its multi-block masks span the full temporal extent of the clip. V-JEPA~2~\citep{assran2025vjepa2} scales this along data, model size, schedule, and resolution, and reports state-of-the-art frozen performance on motion understanding and action anticipation. To use these representations for planning, V-JEPA~2-AC freezes the encoder, applies it per frame, and trains a separate block-causal predictor conditioned on actions and end-effector states on robot interaction data, supporting zero-shot goal-conditioned manipulation through model-predictive control. Other work learns action-conditioned dynamics in pixel space through video generation~\citep{hafner2019dream, bruce2024genie}, at substantially higher planning cost. In each case the temporal model is a separate, action-supervised stage fitted after pretraining; LeVJEPA instead places the temporal constraint in the encoder during pretraining.

\section{Methodology: LeVJEPA}
\label{sec:methodology}

Our goal is to learn visual representations from video with a single network and a single loss, discarding the collapse-prevention heuristics that video joint-embedding methods have so far relied on. LeVJEPA transfers the LeJEPA recipe~\citep{balestriero2025lejepa}, namely the invariance loss paired with SIGReg as the sole mechanism for preventing representation collapse, to video, on top of a V-JEPA-style video transformer backbone~\citep{bardes2024vjepa}.

Given a video, we sample a clip of $16$ frames and construct $V+1$ views of it: one \emph{global} view $x_0$ at full resolution and $V$ \emph{local} views $x_1, \dots, x_V$ obtained by aggressive spatial cropping and photometric augmentation. All views share the identical temporal window and differ only spatially and photometrically. Each view is processed by the same encoder $E_\theta(\cdot)$; a learnable \texttt{[cls]} token provides a clip-level readout, which a small projector $h_\phi(\cdot)$ maps to an embedding $z_v = h_\phi\!\left(E_\theta(x_v)_{\texttt{[cls]}}\right) \in \mathbb{R}^{K}$. This projection is necessary because the final encoder layer applies layer normalization~\citep{ba2016layer}, which constrains the \texttt{[cls]} representation to a sphere and thereby prevents the SIGReg objective from being optimized effectively in the encoder's output space~\citep{kuhn2026levljepaendtoendvisionlanguagepretraining}. The training loss is applied in this $K$-dimensional space; the projector is discarded after pretraining and downstream tasks operate on the encoder's representations.

Joint-embedding methods for video have thus far relied on architectural asymmetries to prevent representation collapse: an exponential-moving-average target encoder, a stop-gradient on the target branch, and a predictor network conditioned on masked-token queries~\citep{bardes2024vjepa, grill2020bootstrap}. While empirically effective, these mechanisms introduce additional networks and schedules whose influence on the learning dynamics is difficult to characterize, and for which no analytical guarantee against collapse is available. In LeVJEPA, collapse is instead excluded by an explicit distributional constraint. A single encoder is trained with the objective
\begin{equation}
    \mathcal{L} \;=\; \mathcal{L}_{\text{inv}} \;+\; \lambda \,
    \mathcal{L}_{\text{SIGReg}},
    \label{eq:total_loss}
\end{equation}
where $\mathcal{L}_{\text{inv}}$ enforces that the embeddings of local views be predictive of the embedding of the global view, and $\mathcal{L}_{\text{SIGReg}}$ constrains the distribution of the embeddings such that collapsed solutions are provably excluded~\citep{balestriero2025lejepa}. The trade-off weight $\lambda$ balances the two terms and constitutes the objective's only hyperparameter; following \citet{balestriero2025lejepa}, it is fixed to $\lambda = 0.02$ in all trainings and is not tuned for any experiment in this paper.

\subsection{Training Objective}
\label{sec:objective}

\subsubsection{Invariance.}
\label{sec:invariance}

The encoder is trained to satisfy the constraint that the embedding of any local view of a clip be predictive of the embedding of the global view of the same clip. As the global view is the only view that remains photometrically unaltered and covers the largest spatial extent, it constitutes the prediction target by construction of the views alone. The invariance term is the mean squared error
\begin{equation}
    \mathcal{L}_{\text{inv}} \;=\; \frac{1}{V+1} \sum_{v=0}^{V}
    \left\lVert z_0 - z_v \right\rVert_2^2 .
    \label{eq:inv}
\end{equation}
Gradients propagate through both variables in Equation~\ref{eq:inv}: the target embedding $z_0$ is produced by the same encoder, in the same forward pass, as the local embeddings; no stop-gradient operation or target network is employed. Minimizing Equation~\ref{eq:inv} in isolation therefore admits a trivial solution in which the encoder outputs a constant embedding regardless of its input. Rather than excluding this solution through the architectural asymmetries discussed above, LeVJEPA excludes it through an explicit constraint on the embedding distribution, described next.

\subsubsection{Regularization.}
\label{sec:sigreg}

SIGReg~\citep{balestriero2025lejepa} constrains the embedding distribution to match an isotropic Gaussian --- the distribution shown by \citet{balestriero2025lejepa} to minimize worst-case downstream probing risk, and from which any collapsed solution, having zero variance along some direction, is maximally distant. By the Cram\'er--Wold theorem, the embeddings match $\mathcal{N}(0, I_K)$ if and only if every one-dimensional projection matches $\mathcal{N}(0, 1)$, reducing the high-dimensional constraint to univariate goodness-of-fit tests. At each step, $M$ directions $a_1, \dots, a_M$ are sampled uniformly on $\mathbb{S}^{K-1}$, and the deviation of each projected batch $\{\langle z_i, a_m \rangle\}_{i=1}^{n}$ from the standard Gaussian is penalized via the Epps--Pulley statistic~\citep{epps1983test},
\begin{equation}
    \mathcal{L}_{\text{SIGReg}} \;=\; \frac{1}{M} \sum_{m=1}^{M}
    \int \Bigl\lvert \tfrac{1}{n} {\textstyle\sum_{i=1}^{n}} e^{\,\mathrm{i}\, t \langle z_i, a_m \rangle} - e^{-t^2/2} \Bigr\rvert^2 \, e^{-t^2/2} \, dt,
    \label{eq:sigreg}
\end{equation}
computed per view and approximated by quadrature. The empirical characteristic function underlying Equation~\ref{eq:sigreg} is bounded with bounded gradients, making the loss robust to outliers and free of whitening or centering operations, and permits distributed evaluation on the full global batch at negligible communication cost; implementation details are provided in Appendix~\ref{app:sigreg}.

\subsection{Architecture}
\label{sec:architecture}

The encoder is a Vision Transformer adapted for video~\citep{dosovitskiy2021vit, arnab2021vivit}. A clip of $16$ frames is tokenized by a convolutional patch embedding of spatial extent $16 \times 16$ and temporal extent $\tau$; by default $\tau = 1$, such that each token corresponds to a patch of a single frame and no temporal aggregation is imposed at the input ($3{,}136$ tokens for a $224^2$ global view, $576$ for a $96^2$ local view). Temporal aggregation ($\tau = 2$)~\citep{arnab2021vivit, tong2022videomae, bardes2024vjepa} is supported and compared against the per-frame default in Section~\ref{sec:design}. During pretraining, a fraction $\rho = 0.95$ of the patch tokens of each view is discarded uniformly at random after patch embedding; the retained tokens constitute a sparse observation of the clip, and this token dropping, which determines the computational cost of pretraining, is analyzed in Section~\ref{sec:design}. A learnable \texttt{[cls]} token prepended to the sequence serves as the clip-level readout, receives the training objective exclusively, and is never dropped; patch tokens receive no direct supervision.

Self-attention follows a block-causal pattern: patch tokens attend bidirectionally within their frame and causally to preceding frames, while the \texttt{[cls]} token attends to all tokens but is not attended to. The representation of a frame is therefore a function of the current and past frames only; this is the default for all results in this paper, and its consequences are examined in Sections~\ref{sec:design}. Positional information is provided by factorized three-dimensional rotary embeddings~\citep{su2024roformer, assran2025vjepa2}, which encode relative position and thereby let the same encoder process both view resolutions without interpolation. The \texttt{[cls]} representation is mapped to the $K$-dimensional embedding space by a small projector shared across views. The trainable architecture comprises the encoder and projector only: no predictor network is instantiated and no target encoder is maintained during training; a Polyak average of the encoder weights is retained solely as the evaluation checkpoint~\citep{busbridge2024scale} and plays no role in the objective. Architectural details are provided in Appendix~\ref{app:architecture}.

\subsection{Pretraining Data and Evaluation Setup}
\label{sec:data_eval}

\paragraph{Pretraining.} Unless stated otherwise, models are pretrained on a class-balanced subsample comprising $20\%$ of K710, the union of the Kinetics-400/600/700 training sets~\citep{kay2017kinetics} with validation overlap removed, following \citet{bardes2024vjepa}. Restricting the default pretraining set keeps the cost of controlled experiments low and, as we show in Section~\ref{sec:comparison}, already suffices for competitive downstream performance.
In addition, we pretrain a ViT-L/16 on the combination of K710, Something-Something-v2~\citep{goyal2017something}, Walking Tours~\citep{venkataramanan2024imagenet}, and the PE Video Dataset released with Perception Encoder~\citep{bolya2026perception}. Walking Tours consists of a small number of hours-long egocentric walking videos; the PE Video Dataset contributes a large corpus of diverse, curated video. Together these sources span curated action-recognition clips, long uncurated egocentric footage, and web-scale general video.

\paragraph{Evaluation.} Pretrained encoders are evaluated frozen, following the attentive-probing protocol of \citet{bardes2024vjepa} exactly: a lightweight cross-attention block with a learnable query pools the encoder's output tokens, and a linear classifier is trained jointly with the probe while the encoder parameters remain fixed. We evaluate action recognition on Kinetics-400~\citep{kay2017kinetics}, motion classification on Something-Something-v2~\citep{goyal2017something}, and object recognition on ImageNet-1K~\citep{russakovsky2015imagenet}, thereby covering appearance-based video understanding, temporal understanding, and static image understanding, respectively. Probing hyperparameters, view sampling at test time, and the adaptation of the video encoder to static images are taken from \citet{bardes2024vjepa} without modification.

Non-linear pooling is appropriate here because the pretraining objective provides no guarantee that the frozen token representations are linearly separable for a given downstream task~\citep{chen2020simple, bardes2024vjepa}; a probe with a learnable attention-based readout evaluates the information content of the representation rather than its linear geometry. On Kinetics-400, whose training set is considerably larger than those of the other benchmarks, we report linear probing on mean-pooled tokens instead, as training the attentive probe at this scale is computationally disproportionate to its purpose; since mean pooling followed by a linear classifier is a strictly weaker adaptation than the attentive probe, the reported Kinetics-400 accuracies constitute a conservative estimate.

\section{What Matters for Efficient Video Pretraining?}
\label{sec:design}

Having specified the method, we examine its design space. The purpose of this analysis is twofold: to determine which choices materially affect the quality of the learned representations, and to characterize the behavior of the objective when components that are standard in video pretraining --- temporal patch aggregation, structured masking, and bidirectional attention --- are removed. Unless stated otherwise, all experiments in this section pretrain a ViT-B/16 on the $20\%$ subsample of K710 described in Section~\ref{sec:data_eval}, and report the top-1 accuracy of a frozen attentive probe on ImageNet-1K, which we found to be the most discriminative single indicator of representation quality among our evaluations.

\subsection{Token Dropping Improves Representations}
\label{sec:design_dropping}

\begin{figure*}[t]
  \centering
  \begin{subfigure}[b]{0.48\textwidth}
    \centering
    \includegraphics[width=\linewidth]{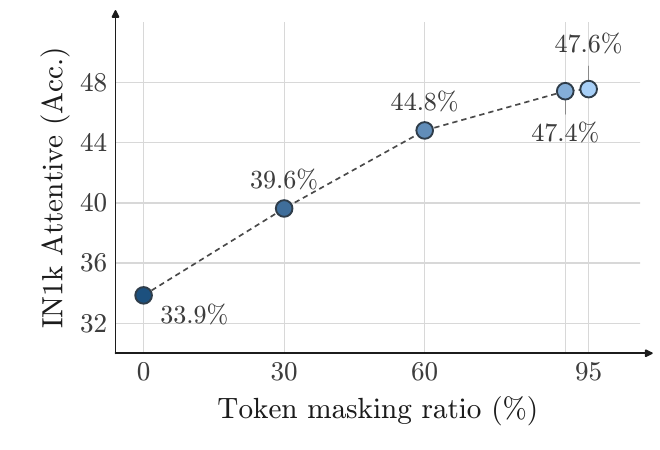}
    \caption{Masking ratio}
    \label{fig:mask_ratio}
  \end{subfigure}
  \hfill
  \begin{subfigure}[b]{0.48\textwidth}
    \centering
    \includegraphics[width=\linewidth]{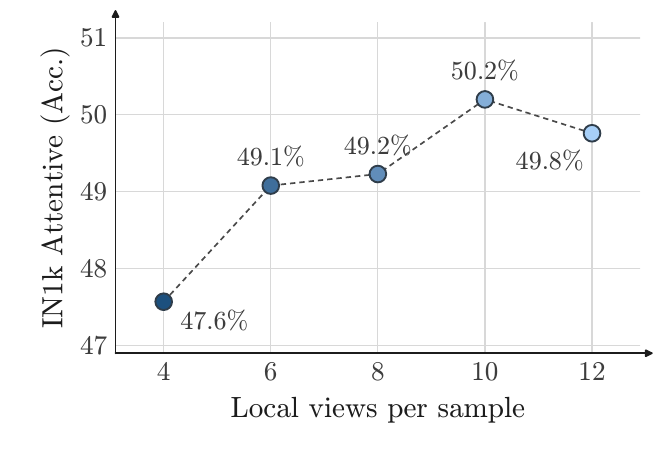}
    \caption{Number of local views}
    \label{fig:local_views}
  \end{subfigure}
  \caption{\textbf{Ablations on view construction.} (\subref{fig:mask_ratio}) Effect of the masking ratio and (\subref{fig:local_views}) effect of the number of local views on ImageNet linear probing accuracy.}
  \label{fig:view_ablations}
\end{figure*}

We first vary the dropping ratio $\rho$ (Figure~\ref{fig:view_ablations}\subref{fig:mask_ratio}). If token dropping constituted merely an approximation adopted for efficiency, downstream accuracy would be expected to degrade as $\rho$ increases. We observe the opposite: ImageNet accuracy increases monotonically with the dropping ratio, from $33.9\%$ when all tokens are processed to $47.6\%$ at $\rho = 0.95$. Token dropping therefore serves a dual role. It reduces the cost of each forward pass by up to a factor of $(1-\rho)^{-1}$ in the feed-forward layers, and it simultaneously acts as a stochastic augmentation that requires the clip-level embedding to be inferable from sparse, randomly located observations of the clip. The comparison between $\rho = 0.9$ and $\rho = 0.95$ is of particular practical relevance: halving the number of processed tokens leaves accuracy unchanged within the observed variability ($47.4\%$ vs.\ $47.6\%$), so the most aggressive dropping ratio considered is simultaneously the most computationally economical configuration.

The benefit of aggressive dropping is, however, not uniform across evaluations. On Something-Something-v2, which primarily probes motion understanding, accuracy declines for dropping ratios beyond $0.3$, in contrast to the monotonic improvement observed on ImageNet. We find that this degradation is mitigated by extending the training schedule: with longer training, higher dropping ratios recover the accuracy of lower ones on Something-Something-v2 while retaining their lower per-iteration cost, such that aggressive dropping remains the more efficient configuration in terms of total pretraining compute. A plausible interpretation is that sparse random observations render motion cues, which depend on correspondences across frames, less frequently recoverable within a single view, and that additional iterations compensate for the reduced per-sample signal; a characterization of this interaction, and dropping schemes that preserve motion information at high sparsity, are left to future work.

The spatial structure of the retained token set is similarly consequential. We compare uniform random dropping against a structured tube variant that retains identical spatial locations in every frame, mirroring the space-time masks employed in masked video modeling~\citep{tong2022videomae, bardes2024vjepa}. Tube dropping reduces accuracy substantially, from $50.7\%$ to $39.6\%$ on ImageNet, and the same ordering holds on Something-Something-v2 ($28.8\%$ against $26.4\%$, under the $\tau = 2$ configuration of Section~\ref{sec:design_tubelets}). We attribute this reversal of the established finding to the difference in objectives. In masked prediction, structured masks are necessary to render the imputation task non-trivial, as randomly distributed masks permit missing content to be interpolated from spatial neighbors. In the present setting, no content is imputed: the retained tokens constitute the encoder's sole observation of the clip. A tube pattern permanently occludes the majority of the scene across all frames, whereas uniform random dropping yields a spatio-temporally distributed sample from which the content of the clip remains identifiable.

% \begin{takeaway}
% Uniform random token dropping is an augmentation, not an approximation: accuracy increases as more tokens are discarded, and the structured masks required by masked prediction become harmful.
% \end{takeaway}

\subsection{Local View Budget}
\label{sec:design_views}

We next vary the number of local views $V$ (Figure~\ref{fig:view_ablations}\subref{fig:local_views}). Accuracy improves consistently from $47.6\%$ at $V = 4$ to $50.2\%$ at $V = 10$, and saturates thereafter ($49.8\%$ at $V = 12$). Each additional local view introduces one further prediction constraint into Equation~\ref{eq:inv} at a marginal cost of approximately $29$ processed tokens, corresponding to less than one fifth of the cost of the global view. We nevertheless retain $V = 4$ for all comparisons in this paper if not specified otherwise, and report this sweep to establish that the results presented in Section~\ref{sec:comparison} do not exhaust the method: additional accuracy is available at a modest increase in pretraining cost by enlarging the view budget alone.

\subsection{Temporal Patch Aggregation Is Not Required}
\label{sec:design_tubelets}

Video transformers conventionally aggregate pairs of consecutive frames at the input ($\tau = 2$), halving the token count prior to the transformer blocks~\citep{arnab2021vivit, tong2022videomae, bardes2024vjepa}. We assess whether this aggregation contributes to representation quality through a comparison in which both the pretraining cost and the evaluation conditions are matched. During pretraining, both configurations process identical $16$-frame clips, and the doubled token count of per-frame patching is offset by a correspondingly halved retention rate ($1-\rho = 0.05$ against $0.1$), such that both encoders retain an equal number of tokens per view; the two dropping ratios are indistinguishable in downstream accuracy (Figure~\ref{fig:view_ablations}\subref{fig:mask_ratio}), so the matching itself does not favor either configuration.

\begin{table}[h]
\centering
\caption{Temporal patch aggregation at a matched token budget. Top-1 accuracy of a frozen attentive probe. Both configurations retain an equal number of tokens per view during pretraining and are evaluated on sequences of $8$ temporal slots.}
\small
\setlength{\tabcolsep}{6pt}
\begin{tabular}{lcc}
\toprule
\textbf{Patch embedding} & \textbf{IN1K} & \textbf{SSv2} \\
\midrule
$\tau = 2$, $\rho = 0.90$ & 47.4 & 28.8 \\
\midrule
\rowcolor{tableblue}
$\tau = 1$, $\rho = 0.95$ & \textbf{50.7} & \textbf{30.4} \\
\bottomrule
\end{tabular}
\label{tab:tubelet}
\end{table}

At evaluation, where no dropping is applied, the sequence length is matched by adjusting the temporal extent of the input to $8$ frames for $\tau = 1$ and $16$ frames for $\tau = 2$, yielding $8$ temporal slots in both cases; for ImageNet, whose static images are repeated along the temporal axis following the protocol of \citet{bardes2024vjepa}, this corresponds to $8$ and $16$ repetitions, respectively. As reported in Table~\ref{tab:tubelet}, the per-frame configuration attains higher accuracy on both benchmarks. The result on Something-Something-v2 is of particular note: temporal aggregation at the input is commonly motivated as a means of capturing short-range motion, yet its removal does not degrade performance on the benchmark most dependent on motion understanding. We conclude that the objective does not depend on temporal aggregation at the input, and that the choice of $\tau$ may consequently be governed by downstream requirements.

% \begin{takeaway}
% Temporal patch aggregation is not needed: per-frame tokenization matches or exceeds it at equal token budget, even on motion-centric evaluation.
% \end{takeaway}

\subsection{Causal Attention Incurs No Accuracy Penalty}
\label{sec:design_causal}

Finally, we compare fully bidirectional attention against the block-causal topology defined in Section~\ref{sec:architecture}, in which tokens attend bidirectionally within a frame and causally across frames. Causal attention constrains the representation of each frame to depend exclusively on the current and preceding frames, a property that bidirectional video encoders lack. Since causal masking removes future tokens from the receptive field of every token, a reduction in representation quality could reasonably be anticipated. Table~\ref{tab:attention} shows that no such reduction occurs: the block-causal encoder matches its bidirectional counterpart ($51.2\%$ against $50.7\%$). Temporal causality is thus obtained at no measurable cost to downstream accuracy, and we adopt block-causal attention as the default for all results reported in this paper.

% \begin{takeaway}
% Block-causal attention matches bidirectional attention, so temporal ordering can be built into the encoder during pretraining rather than fitted afterwards.
% \end{takeaway}

\begin{table}[h]
\centering
\caption{Effect of the attention topology. ImageNet-1K top-1 accuracy of a frozen attentive probe; both configurations use $\tau = 1$, $\rho = 0.95$, $V = 4$, and uniform random dropping.}
\small
\setlength{\tabcolsep}{6pt}
\begin{tabular}{lc}
\toprule
\textbf{Attention} & \textbf{IN1K top-1} \\
\midrule
Bidirectional & 50.7 \\
\midrule
\rowcolor{tableblue}
Block-causal & \textbf{51.2} \\
\bottomrule
\end{tabular}
\label{tab:attention}
\end{table}

\section{Comparison with Prior Work}
\label{sec:comparison}

\subsection{Comparison to Video Models}
\label{sec:comparison_video}

Comparisons between self-supervised video models are frequently confounded by differences in pretraining data, schedule, and compute. To remove these confounds, we retrain all baselines on the identical $20\%$ subsample of K710 used throughout this paper, using their official implementations and recommended hyperparameters, for the same number of epochs ($240$) and at the same effective batch size ($3{,}072$) as our models. All encoders are evaluated with the frozen attentive-probing protocol of \citet{bardes2024vjepa}; since our encoders operate on per-frame tokens, the evaluation sequence length is equalized across methods by adjusting the temporal extent of the probe input, as described in Section~\ref{sec:design_tubelets}, such that every method is probed on the same number of tokens.

\begin{figure*}[t]
  \centering
  \begin{subfigure}[b]{0.495\textwidth}
    \centering
    \includegraphics[width=\linewidth]{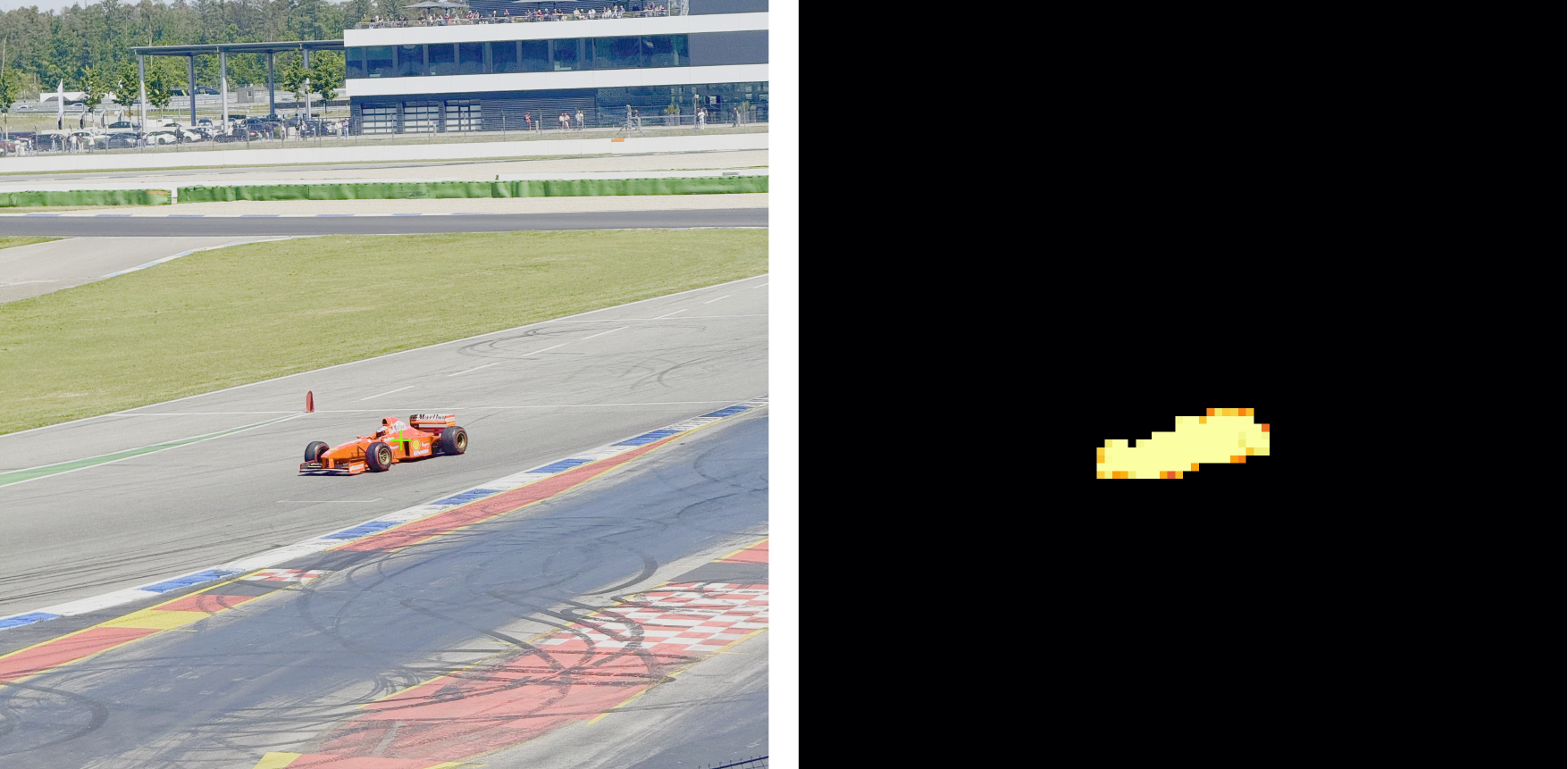}
    \caption{Cosine similarity}
    \label{fig:dense_cossim}
  \end{subfigure}
  \hfill
  \begin{subfigure}[b]{0.48\textwidth}
    \centering
    \includegraphics[width=\linewidth]{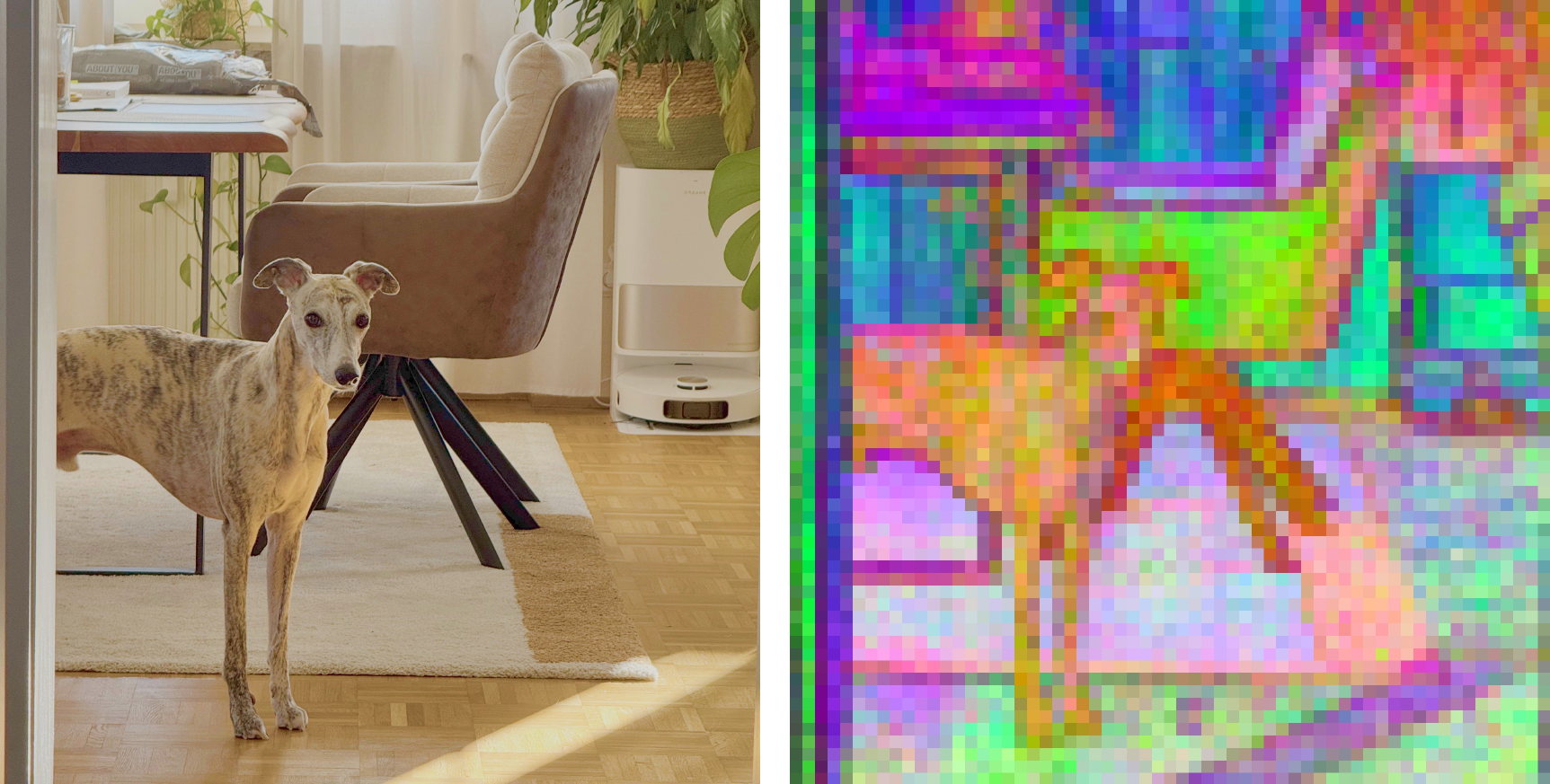}
    \caption{Patch-token PCA}
    \label{fig:dense_pca}
  \end{subfigure}
    \caption{\textbf{Further patch-token visualizations.} (\subref{fig:dense_cossim})~Cosine similarity between a query patch placed on the object and all patch tokens of the image: high similarity is confined sharply to the object, indicating that the token representations are not only semantically organized but spatially precise. (\subref{fig:dense_pca})~Patch-token PCA as in Figure~\ref{fig:pca_comparison}, for a different scene: the decomposition again groups tokens by semantic region, separating the animal from furniture, floor, and background, indicating highly semantic patch tokens.}
      \label{fig:dense_features}
\end{figure*}

Figure~\ref{fig:inet_flops} reports ImageNet accuracy against total pretraining compute for three encoder sizes under this epoch-matched protocol. Across all sizes, LeVJEPA attains accuracy comparable to V-JEPA~2 at a small fraction of the total pretraining compute, with the efficiency advantage ranging from $5.6\times$ at ViT-L to $20.8\times$ at ViT-S. At ViT-B, the two methods are separated by less than one accuracy point while LeVJEPA requires $4.8$ rather than $36.4$ ExaFLOPs; at ViT-L, LeVJEPA is no longer merely comparable but surpasses V-JEPA~2 by $1.9$ points at $5.6\times$ lower cost --- notably, its ViT-L consumes less than half the compute of the V-JEPA~2 ViT-S.

This efficiency follows directly from the mechanisms of Section~\ref{sec:design}: no full-length token sequence is processed at any point during pretraining, and no target-encoder or predictor forward passes are performed. VideoMAEv2 occupies an intermediate position in Figure~\ref{fig:inet_flops}, below both methods in accuracy at a compute cost between them.

% \begin{takeaway}
% Removing collapse-prevention tricks converts directly into compute: LeVJEPA matches V-JEPA~2 at up to $20.8\times$ less pretraining cost, and leads it by $7.6$ ImageNet points at equal FLOPs.
% \end{takeaway}

\begin{table}[h]
\centering
\caption{FLOP-matched comparison on identical pretraining data. ViT-B encoders pretrained on the $20\%$ subsample of K710 at equal total pretraining FLOPs, evaluated frozen; IN1K and SSv2 report attentive-probing top-1 accuracy, K400 reports linear-probing top-1 accuracy.}
\small
\setlength{\tabcolsep}{5pt}
\begin{tabular}{lccc}
\toprule
\textbf{Method} & \textbf{IN1K} & \textbf{SSv2} & \textbf{K400} \\
\midrule
VideoMAEv2 & 53.4 & \textbf{43.6} & 37.4 \\
V-JEPA~2   & 51.6 & 42.5 & 40.7 \\
\midrule
\rowcolor{tableblue}
LeVJEPA    & \textbf{61.0} & 40.4 & \textbf{44.6} \\
\bottomrule
\end{tabular}
\label{tab:flop_matched}
\end{table}

\paragraph{FLOP-matched comparison.} The epoch-matched protocol above holds the number of training samples fixed and lets total compute vary; we now instead hold total pretraining compute fixed. All methods train ViT-B encoders at an equal total FLOP budget; since LeVJEPA processes each sample at a fraction of the baselines cost, the equal budget grants it a correspondingly longer schedule of $1{,}085$ epochs with $V = 10$ local views. Table~\ref{tab:flop_matched} reports the outcome on all three benchmarks. Under equal total compute, LeVJEPA attains the highest ImageNet accuracy by a margin of $7.6$ points and the highest K400 linear-probing accuracy, while being close in the motion based evaluation, reaching within $3.2$ points of the strongest baseline on Something-Something-v2 ($40.4\%$ against $43.6\%$).

\subsection{Comparison to Image Models}
\label{sec:comparison_image}

Image-based self-supervised learning has so far constituted the stronger pretraining paradigm for appearance-centric transfer: video-pretrained encoders have consistently trailed their image-pretrained counterparts on static image benchmarks, such that video pretraining had to be motivated by motion understanding alone~\citep{venkataramanan2024imagenet, bardes2024vjepa}. We revisit this comparison under matched compute by training DINOv2~\citep{oquab2023dinov2} with its official implementation on individual frames drawn from the identical video data ($11.7$M frame samples over $11{,}400$ optimizer steps) at the same total pretraining FLOPs as the $240$-epoch LeVJEPA ViT-B. As reported in Table~\ref{tab:image_comparison}, the image-pretrained encoder retains an advantage of $3.1$ points on ImageNet, while the video-pretrained encoder attains nearly twice its accuracy on Something-Something-v2 ($30.4\%$ against $16.9\%$).

\begin{table}[h]
\centering
\caption{Comparison to image-based pretraining at matched compute. ViT-B encoders pretrained at equal total FLOPs on [identical source data]; top-1 accuracy of frozen attentive probes.}
\small
\setlength{\tabcolsep}{6pt}
\begin{tabular}{lcc}
\toprule
\textbf{Method} & \textbf{IN1K} & \textbf{SSv2} \\
\midrule
DINOv2 & \textbf{53.8} & 16.9 \\
\midrule
\rowcolor{tableblue}
LeVJEPA & 50.7 & \textbf{30.4} \\
\bottomrule
\end{tabular}
\label{tab:image_comparison}
\end{table}

To our knowledge, this is the first FLOP-matched comparison in which video pretraining reaches near-parity with a state-of-the-art image method on appearance-centric evaluation while retaining a decisive advantage on motion-centric evaluation. The implication extends beyond the present method: if the appearance cost of pretraining on video can be reduced to a few points at equal compute then video, which supplies temporal structure that static images cannot, becomes the more FLOP-efficient pretraining substrate for general-purpose visual representations.

% \begin{takeaway}
% At matched compute on identical data, video pretraining nearly reaches image pretraining on appearance while doubling it on motion.
% \end{takeaway}

\subsection{Pretraining on consumer hardware} The low per-sample cost of the method extends to what hardware suffices for pretraining. We train a ViT-Tiny for $12$ hours on a single consumer GPU (RTX~5080, $16$~GB) on eight videos of the Walking Tours dataset~\citep{venkataramanan2024imagenet}, i.e., unlabeled, uncurated egocentric footage totalling approximately $620$k frames, from which the model processes roughly $5$M clips. ImageNet top-1 accuracy of the frozen encoder improves from $8.9\%$ at initialization to $25.2\%$. Beyond the reduced FLOPs, the small memory footprint of sparse token sequences is what makes such training practical: on the same $16$~GB device, LeVJEPA trains at batch size $128$ in under $8$~GB, whereas a V-JEPA configuration with an identically sized encoder saturates the card at batch size $28$. Video pretraining with the present method is thus feasible not only at reduced cluster budgets but on commodity hardware.

\subsection{Scaling the Pretraining Data}
\label{sec:scaling_data}

The comparisons above deliberately restrict pretraining to the $20\%$ subsample of K710 to permit controlled, retrained baselines. We now remove this restriction and ask whether the method continues to improve as the pretraining corpus grows. We pretrain a ViT-L/16 for $100$ epochs on the combined corpus of Section~\ref{sec:data_eval}, comprising K710, Something-Something-v2, Walking Tours, and the PE Video Dataset~\citep{bolya2026perception}. The resulting encoder reaches $69.5\%$ top-1 accuracy on ImageNet-1K and $55.0\%$ on Something-Something-v2 under frozen attentive probing, improving over the ViT-L trained on the $20\%$ subsample (Figure~\ref{fig:inet_flops}) by $9.5$ points on ImageNet within a shorter schedule.  The method thus benefits from additional data without any adjustment of the objective or its single hyperparameter, and the corpus used here remains orders of magnitude below the internet-scale collections of V-JEPA~2~\citep{assran2025vjepa2}, indicating headroom rather than saturation.

\subsection{Emergent Token-Level Structure}
\label{sec:dense_features}

The training objective of LeVJEPA supervises a single clip-level token; the patch tokens receive no loss at any point during pretraining. Figure~\ref{fig:pca_comparison} shows that these unsupervised tokens nevertheless acquire semantically organized representations: the three leading principal components of the patch tokens, visualized for a single image, group by semantic region and separate the object from its surroundings, a degree of visible organization that V-JEPA~2 does not exhibit and that V-JEPA~2.1 obtains through an explicitly introduced auxiliary patch-level objective~\citep{murlabadia2026vjepa21unlockingdense}, both evaluated from their publicly released checkpoints. Figure~\ref{fig:dense_features} extends this picture: the decomposition is consistent across scenes, and the cosine similarity to a query patch is confined sharply to the object, indicating that the representations are spatially precise in addition to semantically organized.

% \begin{takeaway}
% Dense, semantically organized patch representations emerge from clip-level supervision alone without auxiliary patch-level objectives.
% \end{takeaway}

\section{Discussion}
\label{sec:discussion}

This work indicates that self-supervised learning from video can be both simpler and substantially more compute-efficient than previously assumed. A single encoder, trained with an invariance loss and a distributional regularizer whose single hyperparameter is not tuned, attains accuracy comparable to or exceeding that of established video joint-embedding methods at a fraction of their total pretraining compute; the target encoder, predictor network, and associated schedules that these methods employ for stability are not required under the present objective. The analysis of Section~\ref{sec:design} extends this simplification to individual design choices: masking structure, temporal patch aggregation, and attention topology become free parameters that can be selected for efficiency or for downstream requirements, and the resulting configuration, sparse random observation of per-frame tokens under block-causal attention, is simultaneously the cheapest and among the most accurate that we evaluate.

% \textbf{Video as a general-purpose pretraining source.}
The comparison to image-based pretraining suggests a broader implication. Image pretraining has constituted the standard source of general-purpose visual representations, with video regarded as a specialized complement for motion understanding, justified where its additional cost is warranted. Under the present objective, this division is no longer imposed by efficiency: at equal total compute on identical source data, the video-pretrained encoder approaches a DINOv2 baseline on appearance-centric evaluation while attaining nearly twice its accuracy on motion-centric evaluation. Since video strictly contains the appearance information of its constituent frames while additionally supplying temporal structure, these results indicate that video is a viable substrate for general-purpose visual pretraining rather than a specialized one, and that the choice between the two paradigms may increasingly be governed by data availability rather than by computational cost. The data-scaling experiment of Section~\ref{sec:scaling_data} is consistent with this trajectory, with both appearance-centric and motion-centric accuracy improving under a growing corpus; a controlled comparison at internet scale remains an open question.

% \textbf{Encoder carries temporal information.}
Block-causal attention opens a corresponding perspective for temporal modeling. Current systems obtain causal dynamics by fitting a temporal model over the frozen outputs of a pretrained encoder, an arrangement that has proven effective for planning and control~\citep{zhou2025dino, assran2025vjepa2, maes2026leworldmodel} but that separates representation learning from temporal structure. Our results show that the two need not be separated: causality can be established during pretraining, at no measurable cost to downstream accuracy, such that the encoder itself provides per-frame state that respects temporal ordering and extends to incoming frames without re-encoding. This positions a single pretrained encoder as a foundation for streaming perception and autoregressive world modeling. 

% \textbf{Limitations and future work}
Several directions remain open. Motion-centric performance is the axis with the most remaining headroom: aggressive token dropping reduces Something-Something-v2 accuracy at short schedules, and although longer training largely recovers the difference within the same compute budget, dropping schemes that preserve temporal correspondences at high sparsity constitute a natural refinement of the method. Our controlled comparisons are conducted on a restricted corpus at up to ViT-L scale; the behavior of the objective at the model and data scales of recent video foundation models, and the interaction of SIGReg with very large batch and model regimes, remain to be characterized, and the favorable scaling observed in Section~\ref{sec:scaling_data} makes this a promising rather than merely open question. The training signal is applied to a single clip-level token, and although semantically organized patch representations emerge without dense supervision, their sufficiency for dense prediction tasks such as segmentation and tracking has not yet been evaluated. None of these directions requires revisiting the formulation itself: each concerns extending a fixed, simple objective to broader regimes, which we regard as the principal advantage of the approach.

In summary, LeVJEPA demonstrates that a single encoder, a single loss, and one fixed hyperparameter suffice for competitive self-supervised learning from video, and that the resulting formulation is efficient enough to pretrain on commodity hardware, causal by construction, and responsive to growing data. We view these properties as complementary: efficiency determines who can train such models, causality determines what they can be used for, and simplicity determines how reliably the results transfer to new regimes. Together they suggest that video, long the most information-rich but least accessible source of visual supervision, is becoming a practical foundation for general-purpose representation learning.

\paragraph{Acknowledgments}

We gratefully acknowledge support from the hessian.AI Service Center (funded by the Federal Ministry of Research, Technology and Space, BMFTR, grant no. 16IS22091) and the hessian.AI Innovation Lab (funded by the Hessian Ministry for Digital Strategy and Innovation, grant no. S-DIW04/0013/003).

This work was co-funded by the European Union (ERC, TAIPO, 101088594 to F.B.) grant. Views and opinions expressed are those of the authors only and do not necessarily reflect those of the European Union or ERC. Neither the European Union nor the granting authority can be held responsible for them.

\newpage
\clearpage

\bibliography{main}

\clearpage
\appendix

\section{SIGReg Implementation Details}
\label{app:sigreg}

The integral in Equation~\ref{eq:sigreg} is approximated by trapezoidal quadrature over $17$ knots $t \in [0, 3]$, and the statistic is computed separately for the embeddings of each view with $M = 1{,}024$ directions per step. Two properties of the formulation are of practical consequence. First, the empirical characteristic function and its gradient are bounded, so the loss is robust to outliers and requires none of the whitening or centering operations of variance--covariance-based regularizers~\citep{bardes2021vicreg}. Second, the formulation is compatible with distributed training at negligible cost: the empirical characteristic function is averaged across all workers with a single all-reduce operation before the statistic is computed, such that SIGReg is evaluated on the full global batch rather than on per-device shards. The communication overhead is a single $2 \times M \times 17$ tensor, independent of both the batch size and the embedding dimension.

\section{Architectural Details}
\label{app:architecture}

Unless stated otherwise, the encoder is a ViT-B/16. The rotary embeddings partition the dimensions of each attention head into three groups, rotated according to the temporal, vertical, and horizontal coordinates of the token, respectively; no absolute positional embeddings are used. The projector is a two-layer multilayer perceptron composed of a linear layer ($d \to 2{,}048$), batch normalization, a GELU nonlinearity, and a final linear layer ($2{,}048 \to K$) with $K = 256$; it is discarded after pretraining, and all downstream evaluations operate on the encoder's representations. The Polyak average of the encoder weights uses decay $0.9999$, updated every $32$ optimizer steps; it receives no forward passes during training and does not appear in the objective, and is therefore distinct from the exponential-moving-average target encoders employed for collapse prevention in prior work~\citep{grill2020bootstrap, bardes2024vjepa}.

\section{Evaluation Details}
\label{app:eval}

\paragraph{Attentive probing.} The probe follows \citet{bardes2024vjepa} without modification. A single cross-attention layer attends from one learnable query token to all output tokens of the frozen encoder; the attended output is combined with the query through a residual connection, passed through a two-layer MLP with a GELU non-linearity and layer normalization, and classified by a linear layer. The probe and classifier are trained jointly on the downstream training set while all encoder parameters remain fixed; optimization hyperparameters are taken from \citet{bardes2024vjepa}. \paragraph{Linear probing on Kinetics-400.} For Kinetics-400, the output tokens of the frozen encoder are averaged into a single vector and a linear classifier is trained on the pooled representation.

\end{document}